\documentclass[letterpaper]{article} 
\usepackage{aaai25}  
\usepackage{times}  
\usepackage{helvet}  
\usepackage{courier}  
\usepackage[hyphens]{url}  
\usepackage{graphicx} 
\usepackage{natbib}  
\usepackage{caption} 
\usepackage{amsmath}
\usepackage{amssymb}
\usepackage{amsthm}
\usepackage{booktabs}
\usepackage{multirow}
\usepackage{algorithm}
\usepackage{algorithmic}
\usepackage{xcolor}
\newtheorem{theorem}{Theorem}

\usepackage{newfloat}
\usepackage{listings}
\DeclareCaptionStyle{ruled}{labelfont=normalfont,labelsep=colon,strut=off} 
\floatstyle{ruled}
\newfloat{listing}{tb}{lst}{}
\floatname{listing}{Listing}
\title{DFF: Decision-Focused Fine-tuning for \\ Smarter Predict-then-Optimize with Limited Data}
\author {
    Jiaqi Yang \textsuperscript{\rm 1}\equalcontrib,
    Enming Liang \textsuperscript{\rm 2}\equalcontrib,
    Zicheng Su \textsuperscript{\rm 1}\thanks{Corresponding authors.},
    Zhichao Zou \textsuperscript{\rm 3},
    Peng Zhen \textsuperscript{\rm 3},    
    Jiecheng Guo \textsuperscript{\rm 3},\\
    Wanjing Ma \textsuperscript{\rm 1},    
    Kun An \textsuperscript{\rm 1}\footnotemark[2]
}
\affiliations {
    \textsuperscript{\rm 1}{The Key Laboratory of Road and Traffic Engineering of the Ministry of Education, Tongji University, China}\\
    \textsuperscript{\rm 2} Department of Data Science, City University of Hong Kong\\
    \textsuperscript{\rm 3} Didi Chuxing, Beijing, China\\
    2410196@tongji.edu.cn,  
    eliang4-c@my.cityu.edu.hk, 
    suzicheng@tongji.edu.cn,
    zouzhichao@didiglobal.com,
    zhenpeng@didiglobal.com,
    jasonguo@didiglobal.com,
    mawanjing@tongji.edu.cn,
    kunan@tongji.edu.cn
}
\usepackage{bibentry}

\begin{document}

\maketitle

\begin{abstract}
Decision-focused learning (DFL) offers an end-to-end approach to the predict-then-optimize (PO) framework by training predictive models directly on decision loss (DL), enhancing decision-making performance within PO contexts.
However, the implementation of DFL poses distinct challenges. Primarily, DL can result in deviation from the physical significance of the predictions under limited data. Additionally, some predictive models are non-differentiable or black-box, which cannot be adjusted using gradient-based methods.

To tackle the above challenges, we propose a novel framework, Decision-Focused Fine-tuning (DFF), which embeds the DFL module into the PO pipeline via a novel bias correction module. 
DFF is formulated as a constrained optimization problem that maintains the proximity of the DL-enhanced model to the original predictive model within a defined trust region.
We theoretically prove that DFF strictly confines prediction bias within a predetermined upper bound, even with limited datasets, thereby substantially reducing prediction shifts caused by DL under limited data. 
Furthermore, the bias correction module can be integrated into diverse predictive models, enhancing adaptability to a broad range of PO tasks. 
Extensive evaluations on synthetic and real-world datasets, including network flow, portfolio optimization, and resource allocation problems with different predictive models, demonstrate that DFF not only improves decision performance but also adheres to fine-tuning constraints, showcasing robust adaptability across various scenarios.

\end{abstract}

%

\section{Introduction}
Predict-then-Optimize (PO) is a framework that uses machine learning to address decision problems under uncertainty, as the parameters of the optimization problem are likely unknown before making the decision \cite{Bertsimas2020}. This framework operates in two stages: In the first stage, auxiliary features are utilized to predict the unknown parameters; in the second stage, decisions are made based on these predictions. A critical limitation of this two-stage framework is that during the training process of the first stage, the generally used loss functions such as MSE aim to minimize fitting error. However, this may not align with the objective of the final decision task. The new framework proposed to address this issue is Decision-focused Learning (DFL), which has been demonstrated to improve the quality of the final decision by customizing the training of the predictive model based on the decision task \cite{Kotary2021,Mandi2023,Sadana2024}. However, directly replacing the two-stage PO framework with DFL still poses significant challenges. 

\textbf{The first challenge is the convergence issues associated with training from scratch based on decision loss\footnote{In this paper, unless otherwise specified, decision loss refers to loss functions used to measure decision quality, such as regret, surrogate loss, approximated loss, etc.} (DL).} This arises from the highly non-convex nature of DL (e.g., regret), combined with potential discontinuities, leading to substantial computational costs in calculating the gradient of the downstream problem \cite{Elmachtoub2022, Tang2022}. 
A promising solution for this problem is constructing a surrogate loss for DFL. 
In existing works, the design of the surrogate losses is based on Fisher consistency, i.e., the surrogate losses are equivalent with DL when the data is infinite \cite{Elmachtoub2022, Shah2024}. However, in real-world scenarios with limited data, DFL may fail to converge or converge to a sub-optimum. 

\textbf{The second challenge is the significant deviation in predictions induced by directly minimizing DL through training}. As the accuracy of the predictions becomes less controllable, this can lead to a loss of the inherent physical meaning present in the two-stage PO framework. For example, suppose we use a simple linear model with coefficient matrix $\beta$ to fit the dataset $\{\boldsymbol{X}, \boldsymbol{y}\}$, where $\boldsymbol{X}\in\mathbb{R}^{N\times p}$ and $\boldsymbol{y}\in\mathbb{R}^{N\times d}$. First, the PO framework adopts the MSE loss, leading to a closed-form coefficient matrix:
\begin{align}
\beta_{\rm PO} = (\boldsymbol{X}^\top \boldsymbol{X})^{-1} \boldsymbol{X}^\top \boldsymbol{y}
\end{align}

\begin{figure}[h]
    \centering
    \includegraphics[width=1\linewidth]{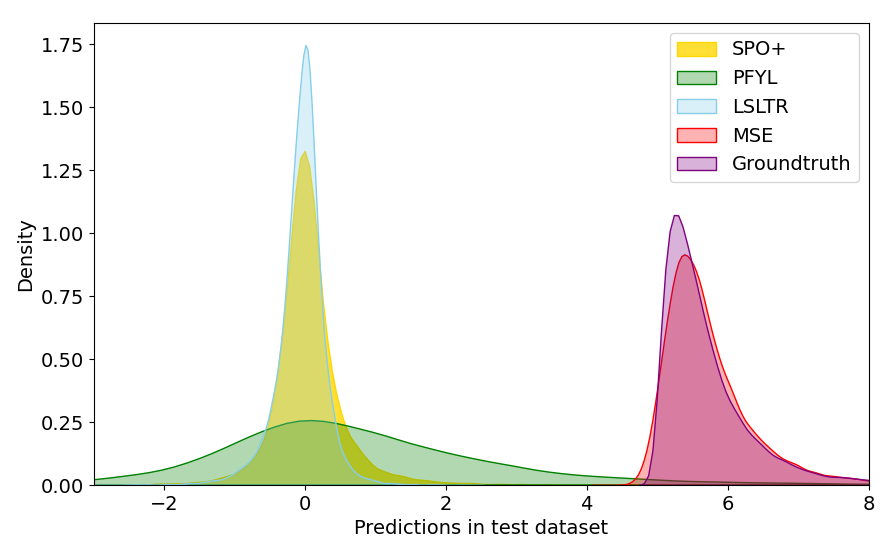}
    \caption{The distribution of predictions generated by different methods for the shortest path problem \cite{Elmachtoub2022} is shown in the figure. It includes the two-stage PO (MSE) and three representative DFL methods, including SPO+ \cite{Elmachtoub2022}, PFYL \cite{Berthet2020}, and LSLTR \cite{Mandi2022}. The results indicate that the predictions from the MSE-based method maintain a distribution similar to the ground truth, whereas the predictions from the DFL models exhibit significant deviations.}
    \label{fig:unexpected-prediction}
\end{figure}

As for the DFL, the predictive model is trained with DL, which is implicitly defined by the decision problem. However, it has been shown that there exists an input-dependent positive semi-definite matrix \( \boldsymbol{Q}_i \) for $i=1,\cdots,N$, leading to a quadratic loss \( L = \frac{1}{N} \sum_{i=1}^N(\boldsymbol{y_i} - \boldsymbol{\hat{y}}_i)^T \boldsymbol{Q}_i (\boldsymbol{y_i} - \boldsymbol{\hat{y}_i}) \), which is Fisher consistent with DL \cite{Shah2024}. Therefore, the coefficient matrix under the surrogate loss is derived from such a generalized least squares problem:
\begin{align}
\beta_{\rm DL} = (\sum_{i=1}^N \boldsymbol{x}_i \boldsymbol{Q}_i\boldsymbol{x}_i^{\top})^{-1}\sum_{i=1}^N \boldsymbol{x}_i \boldsymbol{Q}_i \boldsymbol{y}_i
\end{align}
Since \(\boldsymbol{Q}_i\) is not an identity matrix, DFL with limited data can introduce biases originating from the downstream decision-making objective. 
In other words, the predictions obtained by DFL focus on decision quality, which can result in undesired phenomena, such as multiplicative shifts \cite{Tang2022}  as shown in Figure~\ref{fig:unexpected-prediction} and fail to capture the underlying physical meanings of predictions.


\textbf{The third challenge is that some prediction models are non-differentiable and may not be directly combined with DFL.} 
Current training schemes for DFL primarily focus on first-order gradient-based methods. However, some popular prediction models, such as tree-based models, are not differentiable \cite{breiman2001random}. 
Moreover, in cases where the confidence in the predictive model is low, white-box or semi-black-box simulation models developed based on principles of the physical world can be used for prediction and offline evaluation \cite{Wu2022}. These models can explicitly provide state transitions between variables, offering robust estimates in the presence of many unobserved confounding variables. Nonetheless, it remains unclear how to effectively apply DFL in non-differentiable backbone scenarios.

To address the aforementioned challenges and make DFL more effective, we propose Decision-Focused Fine-Tuning (DFF), which consists of a given backbone predictive model and an additional bias correction layer, as shown in Figure~\ref{fig:network}. The contributions of this paper are as follows:
\begin{itemize}
    \item By utilizing the bias correction layer, DFF can fine-tune the output of any backbone model to produce predictions that are aligned with the decision-making objective.
    \item By explicitly constraining the correction layer, DFF can prevent the fine-tuned output from shifting the inherent physical meaning of the backbone model’s predictions.
    \item We theoretically analyze the performance of DFF on the adherence to fine-tuning constraints and strictly limit the bias in the predictions.
    \item We conduct extensive simulations, including benchmark network flow and portfolio optimization problems and two real-world ride-sourcing subsidy allocation problems. The results show that DFF consistently achieves better decision quality compared to backbone models. 
\end{itemize}

\section{Related Works}
\subsection{Decision Focused Learning}

Since the introduction of OptNet by \citeauthor{Amos2017}, which integrates optimization problems as individual layers within end-to-end trainable deep networks, many works have explored how to better integrate generic optimization problems with machine learning. For convex optimization problems, the gradient can be accurately computed using implicit differentiation methods to guide training \cite{Donti2017, Wilder2019}. However, discrete optimization problems and linear optimization problems often lack gradient information that can guide training, necessitating the development of techniques to obtain approximate gradients \cite{Mandi2023}. These techniques include relaxing and smoothing the optimization problem \cite{Wilder2019}, approximating gradients using random perturbations \cite{Berthet2020, Dalle2022}, and designing surrogate losses \cite{mandi2020interior}. The SPO+ loss function proposed by \citet{Elmachtoub2020} is the first surrogate loss with theoretical guarantees, suitable for optimization problems with any linear objective function. It has been proven to be Fisher consistent with DL. Subsequent research has further explored designing surrogate loss functions for general optimization problems, leading to methods such as LODL \cite{Shah2022}, EGL \cite{Shah2024}, LANCER \cite{zharmagambetov2024landscape}, and TaskMet \cite{Bansal2023}. As for constrained optimization, several studies propose effective methods for addressing scenarios in which uncertain parameters are incorporated into the constraints of optimization problems \cite{hu2023predict+, hu2024two}. Furthermore, thanks to packages like \texttt{CvxpyLayers} \cite{Agrawal2019} and \texttt{PyEPO} \cite{Tang2022}, deploying these complex loss functions now comes with lower engineering costs. Therefore, DFL has the potential to become a modular tool that allows predictive models to be enhanced for specific optimization problems.

\subsection{DFL with Special Prediction Models}

However, DFL faces challenges under some special upstream predictive models, e.g., tree-based model \cite{Elmachtoub2020}, semi-parametric \cite{Wu2022,Zhao2019}, and simulation-based physical model \cite{She2024}. These models are favored under specific settings; for example, tree-based models are preferred in the industry due to their high accuracy and strong interpretability. However, these models are non-differentiable, making it difficult for the gradient of the DL to be backpropagated efficiently.

Existing works have combined tree-based models directly with DFL. \citet{Elmachtoub2020} propose SPO Trees and SPO Forests by modifying the splitting criteria of decision trees, which shows smaller decision regret in linear optimization problems compared to the decision tree and random forest based on MSE. \citet{Butler2023} combine gradient boosting with DFL and propose Dboost, which performs better in convex cone optimization problems. However, numerical experiments by \citet{Butler2023} reveal that the decision quality of Dboost and SPO forest is inferior to gradient boosting based on MSE in certain scenarios, which limits their practicality. To date, there is a lack of an efficient framework to incorporate DFL into general predictive models.

\subsection{Fine-tuning Approach}

Fine-tuning is a crucial technique in deep learning, which involves taking a pre-trained mode and adjusting its parameters to better fit a particular task. This approach allows practitioners to leverage the knowledge embedded in the pre-trained model, significantly speeding up the training process and improving performance on the new task \cite{Zhang2023, Ding2023, Fu2023, Fan2023}.

Under the context of DFL, fine-tuning refers to further adjusting the model’s predictions to reduce the loss associated with the DL. Since the training process of DFL typically requires repeatedly solving the optimization problem, it can reduce the overall training time and improve convergence efficiency. For example, a checkpoint with good prediction performance is obtained by training on MSE, and then further fine-tuning is conducted with DL until convergence \cite{Mandi2020,kotary2022end}. 

\citet{Beichter2024} propose a retraining method that fine-tunes a pre-trained predictive model with weighted MSE and DL and applies it to the dispatchable feeder optimization problem, which resulted in significant performance improvements. 

Nevertheless, existing fine-tuning/retraining can not be applied to general prediction models and there is no performance guarantee of the tuned model when considering the shift under limited data.

\section{Problem Definition}
\subsection*{Predict-then-Optimize}

In the two-stage predict-then-optimize framework, we first train a predictive model $\boldsymbol{M}$ using the MSE loss, with input features $\boldsymbol{x}$ that produce predictions $\boldsymbol{\hat{c}} = \boldsymbol{M}(\boldsymbol{x})$. These predictions serve as the input parameters for the second-stage optimization problem and determine the optimal decision $\boldsymbol{w}^*(\boldsymbol{\hat{c}})$:

\begin{equation}
\boldsymbol{w}^*(\boldsymbol{\hat{c}}) = \arg \min_{\boldsymbol{w}} f(\boldsymbol{w}, \boldsymbol{\hat{c}})
\label{eq:opt_problem}
\end{equation}
\begin{equation}
\text{s.t.} \quad g_j(\boldsymbol{w}) \leq 0, \quad \text{for } \, j \in \{1, 2, \ldots, J\}.
\label{eq:constraints}
\end{equation}
where $f(\cdot)$ represents the objective function of the downstream optimization task, and $g_j(\boldsymbol{w})$ denotes the constraints on the decision variable $\boldsymbol{w}$.
Here, $\boldsymbol{\hat{c}} \in\mathbb{R}^{d}$ represents the unknown parameters, with dimension $d$ typically greater than 1. For each decision problem, our goal is to minimize the decision regret, introduced by \citet{Elmachtoub2022}:
\begin{equation}
\boldsymbol{DR}(\boldsymbol{c}, \boldsymbol{\hat{c}}) = f(\boldsymbol{w}^*(\boldsymbol{\hat{c}}), \boldsymbol{c}) - f(\boldsymbol{w}^*(\boldsymbol{c}), \boldsymbol{c})
\label{eq:decision_regret}
\end{equation}

Given a dataset with \(N\) samples, denoted as \(\mathcal{D} = \{(\boldsymbol{x}_1, \boldsymbol{c}_1), (\boldsymbol{x}_2, \boldsymbol{c}_2), \ldots, (\boldsymbol{x}_N, \boldsymbol{c}_N)\}\), the predictive model $\boldsymbol{M}$ is trained to minimize the average decision regret as follows:
\begin{equation}
\overline{\boldsymbol{DR}} = \frac{1}{N} \sum_{i=1}^{N} \boldsymbol{DR}(\boldsymbol{c}_i,\boldsymbol{M}(\boldsymbol{x}_i))
\label{eq:avg_decision_regret}
\end{equation}

\subsection*{Open Issues for DFL}

As shown in previous research, the DFL method tends to outperform two-stage PO when the predictive model is misspecified or when there are limitations to further improving prediction accuracy \cite{Hu2022, Elmachtoub2023}. However, the application of DFL in such scenarios is also hindered by several limitations.

\subsubsection{Biased Prediction under Limited Data} 

DFL tends to deliver biased predictions, especially under limited data conditions. The reasons are as follows. Firstly, as we mentioned earlier, models trained using DL inherently introduce bias to better align with the decision-making objective. Secondly, existing surrogate loss functions may not satisfy Fisher's consistency under limited data, leading to potential bias compared to DL \cite{Elmachtoub2022}. Lastly, the multiplicative shifts induced by DL can cause predictions to lose their inherent physical meaning \cite{Tang2022}, which is critical in helping analyze downstream decision tasks.

\subsubsection{Non-differentiable Predictive Model}

There are scenarios where designing a suitable predictive model and training it to achieve sufficient accuracy can be challenging. In such cases, an alternative is to utilize simulation-based models. These models can incorporate explicit state transitions based on principles of the physical world and merge prior knowledge of the task. However, it is unclear how to directly apply DFL to these non-differentiable models, and replacing simulation models with DFL-based predictive models is unrealistic. Moreover, since these simulation-based models have been validated as reliable in numerous cases, DFL is expected to enhance decision-making by building on the simulation results rather than simply overturning them. This motivates us to explore how to leverage DFL in these problems while maintaining the existing solution framework and making only minor adjustments to the outputs.

\section{Decision-Focused Fine-tuning}
\begin{figure*}[ht]
\centering
\includegraphics[width=0.9\textwidth]{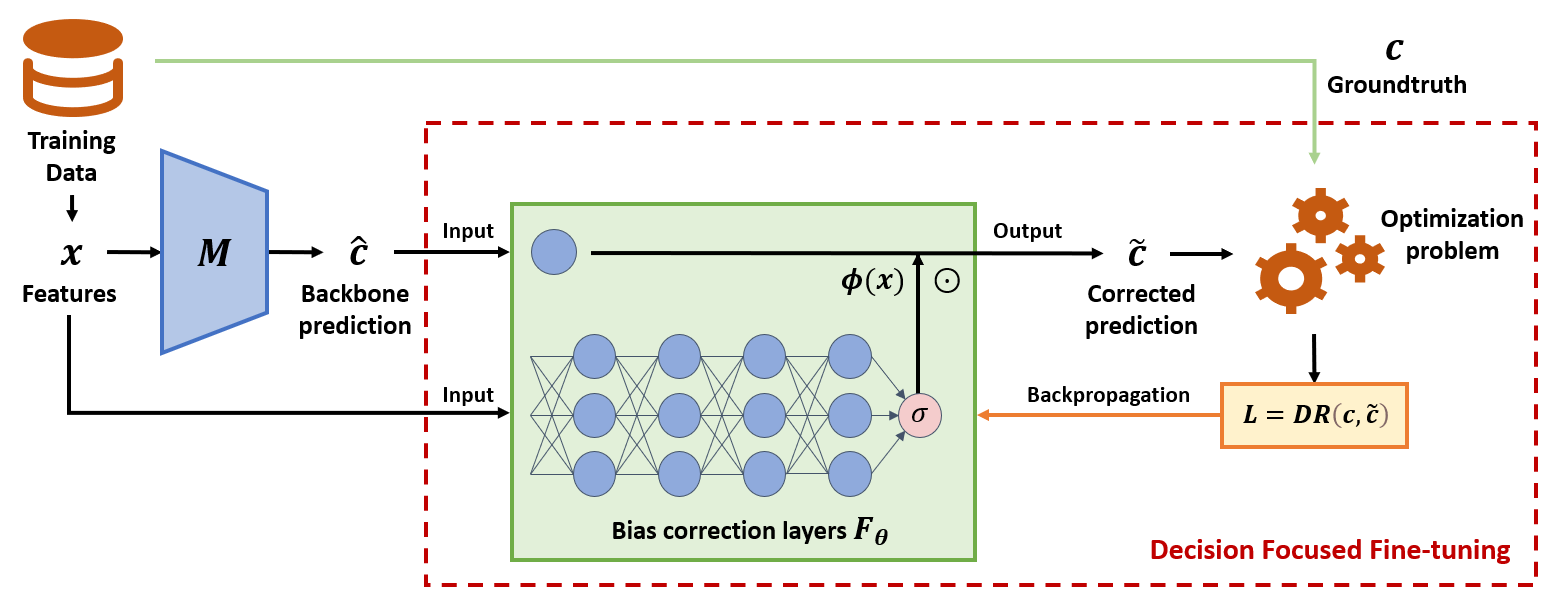} 
\caption{Illustration of the Decision-Focused Fine-tuning Framework.}
\label{fig:network}
\end{figure*}

In this section, we introduce the proposed framework, Decision-Focused Fine-Tuning (DFF), as illustrated in Figure~\ref{fig:network}. We consider a scenario where an upstream predictive model \( \boldsymbol{M} \) has already been established, and its predictions are denoted as \( \boldsymbol{\hat{c}} \). We then design a neural network \( \boldsymbol{F_\theta} \) with a constrained output range, by which \( \boldsymbol{\hat{c}} \) is explicitly represented through a residual connection. All parameters $\theta$ of \( \boldsymbol{F_\theta} \) are optimized to perform a linear transformation on \( \boldsymbol{\hat{c}} \) that minimizes the DL, with the output denoted as \(\boldsymbol{\tilde{c}}\).

\subsection*{Constrained Fine-tuning}

To preserve the advantages of the two-stage framework when applying DFL, our objective is to tune the existing predictive model \( \boldsymbol{M} \) to achieve better decision quality, while ensuring that the adjusted predictions \(\boldsymbol{\tilde{c}}\) remain within a reasonable range from the predictions of the original model, i.e. \( \boldsymbol{\hat{c}} \). This can be formulated as a constrained optimization (CO) problem as follows:

\begin{align}
\label{eq:co}
    \min_{\boldsymbol{F_{\theta}}} &\quad \mathbb{E}_{x,c} \left[\boldsymbol{DR}(\boldsymbol{c},\boldsymbol{F_\theta} (\boldsymbol{x}))\right] \\
    {\rm s.t.} & \quad {\mathbb{D}}(\boldsymbol{F_\theta(x)}, \boldsymbol{M(x)})\le \epsilon  \label{eq:distance}
\end{align}
where $\mathbb{D}$ is a distance metric, e.g., Euclidean distance, in a specific scenario, and $\epsilon$ defines the acceptable range for adjusting the upstream predictive model. This formulation can be interpreted as improving the decision performance by fine-tuning the upstream predictive model \( \boldsymbol{M}\) via \( \boldsymbol{F_\theta} \) within the trust region, particularly under limited data. 

We make the following remarks on the effects of the parameter $\epsilon $:
\begin{itemize}
    \item   When $\epsilon = 0$, no adjustments are made to the upstream predictive model, leading to $\boldsymbol{F_\theta}=\boldsymbol{M}$.
    \item   When $\epsilon \to \infty$, it indicates that $\boldsymbol{F_\theta}$ is trained entirely based on the DL without any restrictions, leading to $\boldsymbol{F_\theta} = \arg \min_{F_{\theta}} \mathbb{E}_{x,c} \left[\boldsymbol{DR}(\boldsymbol{c},\boldsymbol{\tilde{c}}) \right]$.
    \item When $\epsilon \in (0,\infty)$, $\boldsymbol{F_\theta}$ is trained using DL within the trust region of the trained predictive model. The trade-off between these two losses can be problem-specific and defined by the user through the parameter $\epsilon$.
\end{itemize}

We can note that the CO formulation bridges the naive weighted sum of MSE and DL by applying the distance metric $\mathbb{D}$ as a constraint. A similar approach can be found in \citet{Beichter2024}, which incorporates the MSE loss into the objective function \eqref{eq:co} as a Lagrangian relaxation. This can be considered as a special case of our CO formulation. The superiority of the CO formulation lies in its applicability to various types of distance metrics $\mathbb{D}$, allowing for a more efficient adjustment of the trade-off through the parameter $\epsilon$.

Such a trust-region optimization formulation has been successfully applied in various areas such as reinforcement learning \cite{pmlr-v37-schulman15}, online learning \cite{wu2017scalable}, and fine-tuning \cite{ kurutach2018model}. It ensures that the fine-tuned results do not deviate significantly from the upstream predictions, enabling the model to effectively handle decision tasks where data is limited and stability is critical \cite{queeney2021uncertainty}, while further enhancing decision quality. However, the challenges of constrained fine-tuning are twofold: some predictive models are non-differentiable, and enforcing the constraint can be difficult \cite{Mandi2020,kotary2022end,Beichter2024}. We address these challenges by proposing a bias correction layer, as introduced in the following section.

\subsection*{Bias Correction Module}

To tackle the above challenges, we propose a bias correction module \( \boldsymbol{F_\theta} \), motivated by the hyper-network design \cite{Ha2017,She2024}. We set \( \boldsymbol{F_\theta} \) in the following semi-parametric form as:

\begin{equation}
\boldsymbol{F_\theta(x)} =  \boldsymbol{\phi(x)} \odot \boldsymbol{M(x)} + \boldsymbol{b(x)}
\end{equation}

where $\odot$ is the Hadamard product. \( \boldsymbol{\phi(x)} \) and \( \boldsymbol{b(x)} \) represent the input-dependent weights and bias, respectively. Together, they fine-tune the predictive model $\boldsymbol{M}$ to reach an improved decision performance. Notably, \( \boldsymbol{F_\theta} \) does not directly tune the parameters of the predictive model $\boldsymbol{M}$; instead, it learns an input-dependent linear transformation. This approach allows it to apply the DL to general non-differentiable predictive models. Moreover, the distance constraint outlined in Equation \eqref{eq:distance} can be easily enforced by designing the weight and bias modules accordingly.

Specifically, we define the distance metric in the trust-region constraint \eqref{eq:distance} as the percentage error:

\begin{equation}
\begin{split}
\label{eq:range}
\mathbb{D}_i (\tilde{c}_i, \hat{c_i}) &= \left| \frac{\tilde{c}_i - \hat{c_i}}{\hat{c_i}} \right|  \leq \epsilon, \quad  i \in \{1, 2, \ldots, d\}
\end{split}
\end{equation}
Here, \(\epsilon\) represents a pointwise adjustable percentage.
The distance measurement of percentage error is advantageous as it normalizes the predictions into unitless values, thereby avoiding the additional costs of parameter adjustment that may arise due to the differences in the magnitudes of the predictions. 

Given the constraint shown in Equation \eqref{eq:range}, we set $\boldsymbol{b(x)}=0$ and apply an offset-scaled Sigmoid transformation to ensure that the output of \( \boldsymbol{F_\theta(x)} \) strictly falls within the specified range and exhibits center-symmetry, i.e. $\frac{\tilde{c}_i}{\hat{c}_i}\in [1-\epsilon, 1+\epsilon]$. The outputs of $\boldsymbol{\phi(x)}$ can be represented as:
\begin{equation} \label{eq:sigmoid}
   \boldsymbol{\phi(x)} = [(1 - \epsilon) + 2\epsilon \cdot \boldsymbol{\sigma(h(x))}] 
\end{equation}
Here, \( \boldsymbol{h(\mathbf{x})} \) denotes the output from the penultimate layer, and \( \boldsymbol{\sigma(x)} = \frac{1}{1 + e^{-x}} \) denotes the standard Sigmoid function.

It is important to clarify that the percentage error shown in Equation \eqref{eq:distance} is not the only instance of the distance metric, but rather an intuitive and effective example. Further discussion on different types of distance metrics can be found in Appendix A. 

We now present the upper bound on the increment of the RMSE induced by the bias correction layer, and the maximum angular gap between the original predictions \(\boldsymbol{\hat{c}}\) and the corrected predictions \(\boldsymbol{\tilde{c}}\).

\begin{theorem}
\label{thm:theorem1}

For a given \(\epsilon\), the increment in RMSE of \(\boldsymbol{\tilde{c}}\) obtained in Equation \eqref{eq:sigmoid} in the main text has an upper bound such that:
\begin{equation}
    \text{RMSE}(\boldsymbol{\tilde{c}}, \boldsymbol{c}) - \text{RMSE}(\boldsymbol{\hat{c}}, \boldsymbol{c}) \leq \frac{\epsilon}{\sqrt{d}}\| \boldsymbol{\hat{c}}\|_2
\end{equation}
where \(\|\cdot\|_2\) denotes the \(L_2\) norm and $d$ represents the dimension of the vector $\boldsymbol{\hat{c}}$.

Moreover, the cosine similarity between \(\boldsymbol{\tilde{c}}\) and \(\boldsymbol{\hat{c}}\) can be calculated as:
\begin{equation}
    \cos \langle \boldsymbol{\tilde{c}}, \boldsymbol{\hat{c}} \rangle = \frac{\boldsymbol{\tilde{c}} \cdot \boldsymbol{\hat{c}}}{\|\boldsymbol{\tilde{c}}\|_2 \|\boldsymbol{\hat{c}}\|_2} 
\end{equation}
where $\cos \langle \cdot, \cdot \rangle \in [-1,1]$ denotes the similarity of directions between two vectors, with a value of 1 indicating the same direction. The lower bound of $\cos \langle \boldsymbol{\tilde{c}}, \boldsymbol{\hat{c}} \rangle$ is as follows:
\begin{equation}
     \cos \langle \boldsymbol{\tilde{c}}, \boldsymbol{\hat{c}} \rangle \geq
    \sqrt{1 - \epsilon^2}
\end{equation}
\end{theorem}
 
\noindent \textbf{ Proof.} See Appendix A.

Theorm~\ref{thm:theorem1} demonstrates that the point-wise distance metric defined in Equation \eqref{eq:range} can impose a strict bound on the vector-wise distance metrics. These bounds on auxiliary metrics enhance the reliability of predictions derived from by DFF.

\subsection{Loss Function and Training}

After determining the constraint on \( \boldsymbol{F_\theta} \), the training process is transformed into a standard DFL training process, which takes the following loss function:
\begin{equation}
L =    \frac{1}{N} \sum_{i=1}^{N} \boldsymbol{DR}( \boldsymbol{c}_i, \boldsymbol{F_\theta} (\boldsymbol{x}_i))\\
\end{equation}

We need to solve the following core gradient optimization problem:
\begin{equation}
\frac{\partial L}{\partial \theta} = \frac{\partial L}{\partial \boldsymbol{w^*}} \frac{\partial \boldsymbol{w^*}}{\partial \boldsymbol{c}} \frac{\partial \boldsymbol{c}}{\partial \theta}
\end{equation}
where the first two terms involve the gradient of the optimization problem, and the third term is the gradient optimization of the predictive model.

It is worth noting that our framework is general; we only need to select an appropriate gradient calculation method based on the characteristics of the optimization task to compute \( \frac{\partial L}{\partial \theta} \). In this paper, we employ a direct approximation method for calculating \( \frac{\partial L}{\partial \boldsymbol{c}} \) using the SPO+ method because of its superior performance in linear objective function scenarios \cite{Elmachtoub2022}. This method constructs a convex upper bound surrogate loss function \( L = \boldsymbol{DR(c, {\tilde{c}}}) \):

\begin{equation}
    L_{SPO+}= \min_{\boldsymbol{w}} \{(2{\boldsymbol{\tilde{c}}} - \boldsymbol{c})^T \boldsymbol{w} \} + 2 {\boldsymbol{\tilde{c}}}^T \boldsymbol{w^*}\boldsymbol{(c)} - f^*(\boldsymbol{c})
    \label{eq:SPOplus_loss}
\end{equation}

For this loss function, the gradient can be computed using the following formula:
\begin{equation}
\frac{\partial L}{\partial \boldsymbol{c}} \approx \frac{\partial L_{SPO+}}{\partial \boldsymbol{c}} = \boldsymbol{w^*} (2{\boldsymbol{\tilde{c}}} - \boldsymbol{c}) - \boldsymbol{w^*} (\boldsymbol{c})
\end{equation}

\section{Case Study}
We evaluate the effectiveness of the proposed DFF on three two-staged PO problems using various datasets and different predictive models.  Firstly, the DFF is tested on two well-established benchmarks using synthetic data. We then validate the DFF on the resource allocation problem with real-world data from the ride-hailing platform DiDi Chuxing. Lastly, we employ the DFF to adjust the predictions from a non-differentiable simulation model. Note that all experiments are run 10 times with different random seeds and the average results are reported.

\begin{table*}[ht]
\centering
\setlength{\tabcolsep}{1mm}
\begin{tabular}{|l|l|l|l|l|l|l|l|l|}
\hline
\multirow{3}{*}{\textbf{Algorithm}} & \multicolumn{4}{c|}{\textbf{Network Flow Problem}} & \multicolumn{4}{c|}{\textbf{Portfolio Optimization Problem}} \\
\cline{2-9}
 & \multicolumn{2}{c|}{\textbf{Dataset 1}} & \multicolumn{2}{c|}{\textbf{Dataset 2}} & \multicolumn{2}{c|}{\textbf{Dataset 1}} & \multicolumn{2}{c|}{\textbf{Dataset 2}} \\
\cline{2-9}
 & \multicolumn{1}{c|}{\textbf{NDR}} & \multicolumn{1}{c|}{\textbf{MSE}} & \multicolumn{1}{c|}{\textbf{NDR}} & \multicolumn{1}{c|}{\textbf{MSE}} & \multicolumn{1}{c|}{\textbf{NDR}} & \multicolumn{1}{c|}{\textbf{MSE}} & \multicolumn{1}{c|}{\textbf{NDR}} & \multicolumn{1}{c|}{\textbf{MSE}} \\
\hline
Random Forest & 6.24\% & 4.43$\times 10^{-1}$ & 8.70\% & 5.75$\times 10^{-1}$ & 23.21\% & 4.60$\times 10^{-1}$ & 35.69\% & 6.92$\times 10^{-1}$ \\
SPO Forest & 5.99 \% (\checkmark)& 5.19$\times 10^{-1}$ & 8.49\% (\checkmark)& 6.44$\times 10^{-1}$ & 22.50\% (\checkmark)& 4.76$\times 10^{-1}$   & 34.50\% (\checkmark)& 7.03$\times 10^{-1}$ \\
\hline
Boost & 5.26\% & 3.69$\times 10^{-1}$ & 5.75\% & 3.85$\times 10^{-1}$ & 16.44\% & 3.26$\times 10^{-1}$ & 16.38\% & 3.47$\times 10^{-1}$ \\
Dboost & 5.86\% & 5.78$\times 10^{-1}$ & 8.32\% & 6.83$\times 10^{-1}$ & 16.56\% & 3.88$\times 10^{-1}$ & 17.87\% & 4.78$\times 10^{-1}$ \\
\hline
NN-MSE & 5.06\% & 3.37$\times 10^{-1}$ & 9.44\% & 6.08$\times 10^{-1}$ & \textbf{14.46\%} & 2.87$\times 10^{-1}$ & 21.98\% & 4.49$\times 10^{-1}$ \\
NN-SPO+ & 4.63\% (\checkmark)& 1.92 & 9.43\%(\checkmark) & 1.80 & 14.74\% & 4.98$\times 10^{-1}$ & 23.55\% & 7.08$\times 10^{-1}$\\
\hline
2-fold Boost & 4.49\% & 3.03$\times 10^{-1}$ & 4.88\% & 3.23$\times 10^{-1}$ & 16.08\% & 3.18$\times 10^{-1}$ & 15.92\% & 3.39$\times 10^{-1}$ \\
\textbf{DFF (ours)} & \textbf{4.39\%} (\checkmark)& 3.11$\times 10^{-1}$ & \textbf{4.81\%} (\checkmark)& 3.26$\times 10^{-1}$ & 15.97\% (\checkmark)& 3.36$\times 10^{-1}$ & \textbf{15.91\%} (\checkmark)& 3.41$\times 10^{-1}$\\
\hline
\end{tabular}
\caption{NDR and MSE of benchmarking methods on two well-established PO problems}
\label{tab:case1}
\end{table*}

\subsection{Benchmarks with synthetic data}

\subsubsection{Problem Settings and Dataset}
In line with \citet{Elmachtoub2022} and \citet{Butler2023}, the proposed DFF framework is first evaluated on two well-established optimization problems: the network flow problem and the portfolio optimization problem. The formulations for these benchmarks are illustrated in Appendix B. Specifically, both problems aim to minimize a linear objective function, while the portfolio optimization problem involves a nonlinear constraint. To conduct a fair comparison among different predictive models, we generate two distinct datasets using different mechanisms, including polynomial functions, periodic functions, and piecewise functions with different coefficients. The details of the synthetic data are presented in Appendix B.

\subsubsection{Baseline}
In this paper, we compare the DFF framework with three state-of-the-art DFL methods, as well as their original predictive models trained with the MSE loss. The following are brief introductions to the baseline models:
\begin{enumerate}
    \item \textbf{Random Forest}: Random forest trained with MSE.
    \item \textbf{SPO Forest} \cite{Elmachtoub2020}: Random forest trained with the decision regret shown in Equation \eqref{eq:decision_regret} using zero-order gradient.
    \item \textbf{Boost}: Gradient boosting tree trained with MSE.
    \item \textbf{Dboost} \cite{Butler2023}: Gradient boosting tree trained with the decision regret in Equation \eqref{eq:decision_regret} using the fixed-point argmin differentiation.
    \item \textbf{NN-MSE}: Neural network trained with MSE.
    \item \textbf{NN-SPO+} \cite{Elmachtoub2022}: Neural network trained with the SPO+ loss shown in Equation \eqref{eq:SPOplus_loss}.
    \item \textbf{2-fold Boost}: Gradient boosting tree based on MSE with 2-fold cross-validation.
    \item \textbf{DFF (ours)}: Fine-tuning the \textbf{2-fold Boost} with the SPO+ loss shown in Equation \eqref{eq:SPOplus_loss}.
\end{enumerate}
The baselines are evaluated with the normalized decision regret (NDR) \cite{Tang2022}, defined as follows:

\begin{equation}
    \boldsymbol{NDR}=
    \frac{\sum_{i=1}^{N} f(\boldsymbol{w}^*(\boldsymbol{\hat{c_i}}), \boldsymbol{c_i}) - f(\boldsymbol{w}^*(\boldsymbol{c_i}), \boldsymbol{c_i})}{\sum_{i=1}^{N} |f(\boldsymbol{w}^*(\boldsymbol{c_i}), \boldsymbol{c_i})|}
\end{equation}

\subsubsection{Hyperparameters and Training Process} In this paper, all tree-based models are trained with a maximum depth of 2 and no more than 100 trees. In particular, the Random Forest model samples the training data with a 50\% sampling rate. As for the neural network, it consists of 3 layers with 32 neurons on each layer, and the ReLU activation function is used. The parameter of constrained distance $\epsilon$ is set to 0.5 for the DFF. Since the Boost model produces the lowest MSE error among the prediction models, we use it as the backbone model to fine-tune its predictions with the SPO+ loss, as shown in Equation \eqref{eq:SPOplus_loss}. To avoid self-fitting and make full use of data, we split the training set into two disjoint datasets, which is in line with the 2-fold cross-fitting method proposed by \citet{Chernozhukov2018}. In Appendix C, we demonstrate that the proposed DFF module also enhances the decision performance of other predictive models.

\subsubsection{Results} The normalized decision regret and MSE of different methods are summarized in Table~ref{tab:case1}. The checkmark `$\checkmark$' represents that the predictive model trained using the DL can improve the NDR of its counterpart trained with MSE. Firstly, we can note that the proposed DFF can improve its base model in both problems across all the datasets. This demonstrates that DFF can effectively enhance the downstream optimization task by adjusting the backbone predictive model within the trust region, even in the presence of nonlinear constraints. Moreover, the DFF method achieves the lowest normalized regrets in all settings, except for the portfolio optimization problem with Datasets 1. This is because the base model, 2-fold Boost, is less efficient than NN-MSE in this scenario, and the fine-tuning based on DFF is constrained by the trust region.

\subsection{Resource allocation problem with real-world data}
We further validate the DFF framework with the resource allocation problem for the ride-hailing platform DiDi Chuxing. The key task is to allocate the subsidy budget between different cities based on the predicted subsidy conversion rate, thereby maximizing the platform's revenue. The detailed formulation is presented in Appendix D.

In practice, the platform employs Extreme Gradient Boosting (XGBoost) as the predictive model, as it offers sufficient predictions, high robustness, and ease of implementation. However, it is challenging to apply DFL to XGBoost for two reasons. On the one hand, it requires calculating the second-order derivative of the loss function to train XGBosst, for which there is currently no corresponding method in existing DFL frameworks. On the other hand, there is limited data available for training such a model, as only the historical market data within three months can be used to make a timely prediction.

\begin{table}[ht]
\centering
\begin{tabular}{ccc}
\toprule
Method               & NDR  & MSE  \\ \hline
\multicolumn{3}{c}{Fine-tuning the XGBoost model}          \\ \hline
XGBoost              & 2.54\%             & 0.45    \\
NN-SPO+              & 2.40\%             & 2.04    \\
\textbf{DFF(ours)}   & \textbf{2.39\%}    & 0.49    \\ \hline
\multicolumn{3}{c}{Fine-tuning the simulation model} \\ \hline
Average allocation   & 3.45\%             & /    \\
Simulation model     & 2.79\%             & 2.59 $\times 10^{-2}$   \\
\textbf{DFF(ours)}   & \textbf{2.11\%}    & 2.61 $\times 10^{-2}$    \\ 
\bottomrule
\end{tabular}
\caption{Normalized regret and MSE across different methods in resource allocation problem.}
\label{tab:case_2}
\end{table}

\begin{figure}
    \centering
    \includegraphics[width=1\linewidth]{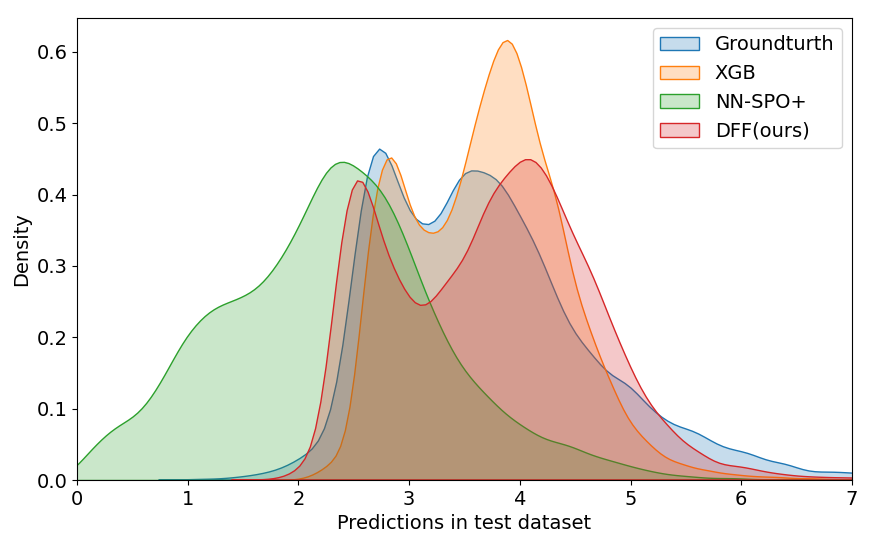}
    \caption{Distribution of predictions with different methods}
    \label{fig:case_2}
\end{figure}

We fine-tune the XGBoost model using DFF with market data from 102 consecutive days from Didi Chuxing and subsequently allocate the subsidy budget across 105 cities. The results are summarized in the first part of Table~\ref{tab:case_2}. The proposed DFF framework outperforms the original prediction backbone XGBoost and the NN-SPO+ method in terms of normalized regret. Notably, the MSE of the DFF framework is significantly lower than that of the NN-SPO+ method, comparable to the XGBoost model which is specifically trained to minimize the MSE loss. Moreover, Figure~\ref{fig:case_2} depicts the distribution of predicted values from different models, which represents the subsidy conversion rate that has a crucial physical meaning. 
Notably, the ground truth exhibits a significant pattern of bimodal distribution. However, the predicted values from the NN-SPO+ method show a considerable deviation, presenting an unimodal distribution. This is consistent with the findings in \citet{Tang2022} that direct training with DL can induce a multiplicative shift in the predicted values. In contrast, the predictions from the DFF method closely align with the ground truth in both magnitude and distribution shape.

\subsection{Fine-tuning Non-differentiable Simulation Model}

In the context of the resource allocation problem, the ride-hailing platform DiDi Chuxing has developed a simulation model to derive the subsidy conversion rate, enabling the analysis of domain knowledge in the ride-hailing market \cite{gao2024evaluation}. However, the downstream DL cannot be directly backpropagated to adjust the simulation model due to its non-differentiable nature. To address this, we apply the proposed DFF module to fine-tune the simulation model based on the DL. Due to the absence of DFL methods for adjusting the simulation model, we incorporate a rule-based average allocation strategy as a baseline. 

\begin{figure}
    \centering
    \includegraphics[width=1\linewidth]{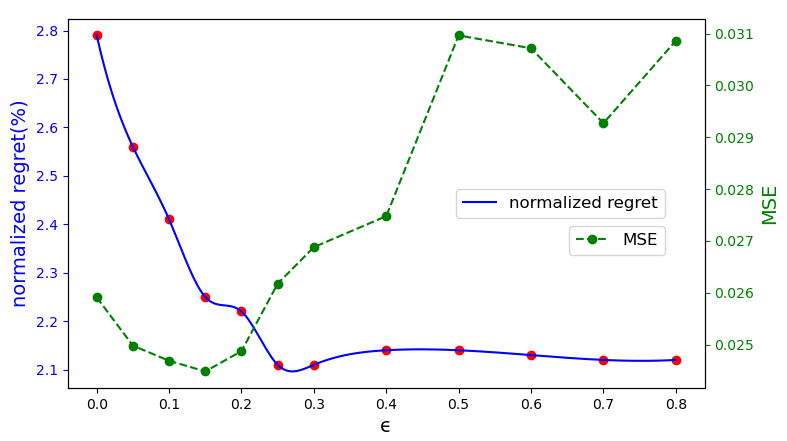}
    \caption{Sensitivity analysis of parameter $\epsilon$ on the decision loss and MSE loss}
    \label{fig:sensitivity}
\end{figure}

The summarized results are presented in the lower part of Table~\ref{tab:case_2}. Notably, by fine-tuning the simulation model with the DFF module, the normalized decision regret is reduced by 24.37\% compared to the original simulation model, while the MSE loss remains at a similar level. This again demonstrates the effectiveness of the constrained fine-tuning design. Furthermore, we conduct a sensitivity analysis on the parameter $\epsilon$ that confines the region in which the predictive model can be adjusted. As shown in Table~\ref{fig:sensitivity}, we observe that when $\epsilon > 0.2$, the normalized decision regret stabilizes at a low level. In contrast, the MSE loss increases with a larger $\epsilon$. Therefore, we can conclude that even by restricting the fine-tuning module within a tight region with a small value of $\epsilon$, the DL can still be improved while maintaining the MSE loss to prevent prediction shift.

\section{Conclusion}

In scenarios with limited data for PO problems, the reliability of prediction results and the quality of final decisions are crucial. Existing DFL approaches can effectively improve the decision quality, but they may introduce significant bias in predictions, which increases the risk of decision-making from a different perspective. To address this issue, this paper introduces DFF, which employs a unique training mechanism that maximizes decision quality while ensuring that fine-tuning constraints are satisfied. The method is applicable to any upstream prediction model, including non-differentiable models. Empirical evaluations across diverse scenarios including synthetic data and real-world resource allocation problems demonstrate that DFF outperforms traditional models and existing DFL approaches. 

Due to the decoupling between DFF and the upstream model, DFF has advantages in multi-task learning and multi-objective optimization scenarios. Furthermore, its constraint fine-tuning design allows it to better handle downstream problems with uncertain parameters in constraints. These are the key focus areas of our future work.

\section{Acknowledgments}
Financial supports from the National Natural Science Foundation of China (No. 52302411, No. 52472349, No. 72361137005, No. 52131204) and CCF-DiDi GAIA Collaborative Research Funds (No. 202310) are gratefully acknowledged.

\bibliography{main}

\end{document}